\documentclass{ifacconf}

\usepackage{graphicx}      %
\usepackage{natbib}        %
\usepackage{amsmath}
\usepackage{amssymb}
\usepackage{bm}
\usepackage{mathtools}
\usepackage{siunitx}
\usepackage{glossaries}

\usepackage{algorithm}
\usepackage{algpseudocode}
\usepackage{lipsum}

\usepackage{subcaption}
\usepackage{subfig}

\usepackage[short,nocomma]{optidef}

\usepackage{xstring}

\usepackage{tikz}
\usepackage{circuitikz}
\usepackage{graphicx}
\usetikzlibrary{positioning} 
\usetikzlibrary{fit}
\usetikzlibrary{calc}
\usetikzlibrary{spy}
\usetikzlibrary{plotmarks}
\usetikzlibrary{arrows.meta}
\usetikzlibrary{shapes.misc}
\usepackage{pgfplots}
\pgfplotsset{compat=newest}
\usepgfplotslibrary{patchplots}

\usepackage{pdfpages}

\usepackage{xcolor}
\newcommand{\todo}[1]{%
  \textcolor{red}{\textbf{%
    TODO%
    \if\relax\detokenize{#1}\relax
    \else:\ #1%
    \fi
  }}%
}

\begin{document}
\begin{frontmatter}

\title{High-Dimensional Surrogate Modeling for Closed-Loop Learning of Neural-Network-Parameterized Model Predictive Control}

\author[First]{Sebastian Hirt}
\author[First]{Valentinus Suwanto}
\author[First]{Hendrik Alsmeier}
\author[First]{Maik Pfefferkorn}
\author[First]{Rolf Findeisen} 

\address[First]{Control and Cyber-Physical Systems Laboratory, \\ Technical University of Darmstadt, Darmstadt, Germany, \\ \{sebastian.hirt, hendrik.alsmeier,
maik.pfefferkorn, rolf.findeisen\}@iat.tu-darmstadt.de.}

\begin{abstract}
Learning controller parameters from closed-loop data has been shown to improve closed-loop performance. Bayesian optimization, a widely used black-box and sample-efficient learning method, constructs a probabilistic surrogate of the closed-loop performance from few experiments and uses it to select informative controller parameters. However, it typically struggles with dense high-dimensional controller parameterizations, as they may appear, for example, in tuning model predictive controllers, because standard surrogate models fail to capture the structure of such spaces.
This work suggests that the use of Bayesian neural networks as surrogate models may help to mitigate this limitation. Through a comparison between Gaussian processes with Matérn kernels, finite-width Bayesian neural networks, and infinite-width Bayesian neural networks on a cart–pole task, we find that Bayesian neural network surrogate models achieve faster and more reliable convergence of the closed-loop cost and enable successful optimization of parameterizations with hundreds of dimensions. Infinite-width Bayesian neural networks also maintain performance in settings with more than one thousand parameters, whereas Matérn-kernel Gaussian processes rapidly lose effectiveness.
These results indicate that Bayesian neural network surrogate models may be suitable for learning dense high-dimensional controller parameterizations and offer practical guidance for selecting surrogate models in learning-based controller design.
\end{abstract}


\end{frontmatter} 

\section{Introduction}

Tuning control parameters to achieve optimal control performance is a challenging task, as often the controller parameters are decoupled from the control goal. This is especially true for model predictive control (MPC), which is by now a widely accepted methodology for controlling complex dynamical systems under constraints \citep{rawlings2017model,schwenzer2021review,lucia2016predictive,Findeisen2025}, as its performance critically depends on the design and parameters of the cost function. Identifying parameters that yield high closed-loop performance is difficult, particularly when classical quadratic cost structures are insufficient to exploit the full performance potential of the plant.

Learning-based controller tuning has therefore gained increasing interest, where controller parameters are directly optimized from data based on their effect on closed-loop behavior \citep{piga2019performance,makrygiorgos2022performanceorienteda,kordabad2023reinforcement}. Among the various approaches, Bayesian optimization has emerged as a powerful and sample-efficient tool for global learning of controller parameters \citep{paulson2023tutorial,hirt2025learning,hirt2024safe}. A key limitation of Bayesian optimization, however, lies in its sensitivity to the dimensionality of the parameter search space: conventional Gaussian process surrogate models degrade significantly in scalability and predictive performance as dimensionality grows.

This challenge becomes particularly relevant when MPC is extended with dense, high-dimensional parameterizations, such as neural-network-augmented cost functions \citep{seel2022convex,hirt2024safe,hirt2024stabilityinformed}, or more generally any parameterization where every parameter influences the closed-loop behavior in a significant way. In contrast to common high-dimensional Bayesian optimization benchmarks, these problems lack latent low-dimensional structure or sparsity: many existing methods exploit additive structure, random embeddings, or compressed latent representations \citep{eriksson2021high-dimensional,moss2025return}, none of which apply here. Recent advances in surrogate modeling based on Bayesian neural networks and infinite-width Bayesian neural networks have not yet been utilized for closed-loop learning of controller parameters, even though their computational scaling makes them attractive candidates for large high-dimensional spaces \citep{li2024study}.

This work examines the use of Bayesian neural networks and infinite-width Bayesian neural networks as scalable probabilistic surrogates for closed-loop learning of MPC parameters in dense high-dimensional spaces. Specifically, we evaluate Bayesian optimization performance in terms of best cost convergence when learning neural-network-parameterized MPC cost functions with more than $1000$ parameters. As a comparison, we consider Gaussian processes with Matérn kernels, which represent the standard surrogate baseline widely used in Bayesian optimization, against Bayesian neural networks and infinite-width Bayesian neural network surrogates. The latter two constitute our contribution toward scalable modeling in dense high-dimensional MPC parameter spaces. The main contribution of this work is to show that Bayesian neural network surrogate models, in particular infinite-width variants, may offer a scalable approach for learning dense high-dimensional model predictive control parameterizations where conventional Gaussian process models become ineffective. Our results provide practical insight into surrogate model selection for learning-based MPC under high-dimensional parameterizations.

Our simulation results on closed-loop learning of MPC parameters show that Bayesian-neural-network-based surrogates yield faster convergence of the best closed-loop cost than Matérn kernels as the number of cost parameters increases, and that infinite-width Bayesian neural networks retain strong performance even beyond 1000 parameters. These findings indicate that Bayesian-neural-network-based surrogates may reliably scale to dense high-dimensional MPC design problems, where conventional Gaussian process models become inefficient, and they provide practical guidance for surrogate model selection in learning-based MPC.

The remainder of this work is structured as follows. In Section~\ref{sec:fundamentals}, we introduce the problem setup, the parameterized MPC formulation, and the probabilistic surrogate models used within Bayesian optimization, as well as the optimization procedure itself. Section~\ref{sec:nn_param} details the dense high-dimensional neural network cost parameterization. Section~\ref{sec:simulation} presents a simulation study on a cart–pole system and compares surrogate performance in terms of best cost convergence. We conclude in Section~\ref{sec:conclusion} and provide directions for future research.
\section{Problem Formulation and Fundamentals}
\label{sec:fundamentals}

We consider a nonlinear discrete-time system described by
\begin{equation}
\label{eq:system}
    x_{k+1} = f(x_k, u_k),
\end{equation}
where $x_k \in \mathbb{R}^{n_\mathrm{x}}$ denotes the system state, 
$u_k \in \mathbb{R}^{n_\mathrm{u}}$ the control input, and 
$f: \mathbb{R}^{n_\mathrm{x}} \times \mathbb{R}^{n_\mathrm{u}} 
\to \mathbb{R}^{n_\mathrm{x}}$ captures the (nonlinear) dynamics.

A model predictive controller (MPC) is used to control the system toward a desired state $(x_\mathrm{d}, u_\mathrm{d})$. The resulting closed-loop behavior depends sensitively on the MPC design choices, in particular on the MPC cost function. To systematically shape the closed-loop behavior and enable data-driven performance tuning, we introduce a parameterization of the MPC design (e.g., cost weights, terminal ingredients), represented by $\theta \in \Theta \subseteq \mathbb{R}^{n_\theta}$.
Consequently, for any $\theta$, the resulting controller induces a closed-loop trajectory $\{x_k(\theta), u_k(\theta)\}_{k=0}^{M}$ over a finite evaluation horizon $M \in \mathbb{N}$. The associated closed-loop cost is given by $G : \Theta \to \mathbb{R}$ and the learning objective is to find MPC parameters that maximize the closed-loop performance
\begin{equation}
\label{eq:learning_problem}
    \theta^* = \arg \max_{\theta \in \Theta} -G(\theta).
\end{equation}

Direct evaluation of $G(\theta)$ requires closed-loop interaction with 
the real plant or high-fidelity simulations and is typically 
expensive, noisy, and not available in closed form. 
Therefore, we adopt Bayesian optimization (BO) as a 
data-efficient black-box optimization method to identify $\theta^*$ 
based solely on performance data obtained from closed-loop executions.

\subsection{Parameterized Model Predictive Control}
We employ a model predictive control formulation parameterized by 
$n_\theta \in \mathbb{N}$ parameters 
$\theta \in \Theta \subset \mathbb{R}^{n_\theta}$.
At each time step $k$, for a given parameter vector $\theta$, 
the controller computes an optimal input sequence by solving the optimal control problem given by
\begin{mini!}
    {\mathbf{\hat{u}}_k}{\left\{ \sum_{i=0}^{N-1} l_\theta({\hat x}_{i \mid k}, {\hat u}_{i \mid k}) + E_\theta({\hat x}_{N \mid k}) \! \right\}\label{eqn:mpc_ocp_cost}}{\label{eqn:mpc_ocp}}{}
    \addConstraint{\forall i}{\in \{0, 1, \dots, N-1\}: \notag}{}
    \addConstraint{}{\hat x_{i+1\mid k} = \hat f_\theta(\hat x_{i\mid k}, \hat u_{i\mid k}), \ \hat x_{0 \mid k} = x_k,}{\label{eqn:mpc_ocp_model}}
    \addConstraint{}{\hat x_{i \mid k} \in \mathcal{X}_\theta, \ {\hat u}_{i \mid k} \in \mathcal{U}, \ \hat x_{N \mid k} \in \mathcal{E}_\theta.}{\label{eqn:mpc_ocp_constraints}}
\end{mini!}
Here, $\hat{x}_{i\mid k}$ and $\hat{u}_{i\mid k}$ denote the predicted state and input $i$ steps ahead from time $k$, and $x_k$ is the current state measurement. The prediction model is given by $\hat f_\theta : \mathbb{R}^{n_\mathrm{x}} \times \mathbb{R}^{n_\mathrm{u}} \to \mathbb{R}^{n_\mathrm{x}}$, the length of the prediction horizon is $N \in \mathbb{N}$, and the parameterized stage and terminal cost functions are defined as $l_\theta : \mathbb{R}^{n_\mathrm{x}} \times \mathbb{R}^{n_\mathrm{u}} \to \mathbb{R}$ and $E_\theta : \mathbb{R}^{n_\mathrm{x}} \to \mathbb{R}$, respectively. The constraints impose admissible sets for states, inputs, and terminal states, denoted by $\mathcal{X}_\theta \subseteq \mathbb{R}^{n_\mathrm{x}}$, $\mathcal{U} \subseteq \mathbb{R}^{n_\mathrm{u}}$, and $\mathcal{E}_\theta \subseteq \mathbb{R}^{n_\mathrm{x}}$.

Solving \eqref{eqn:mpc_ocp} yields the optimal input sequence
$\mathbf{\hat{u}}_k^*(x_k; \theta)
=[\hat u_{0 \mid k}^*(x_k; \theta),\dots,\hat u_{N-1 \mid k}^*(x_k; \theta)]$.
Only its first element is applied to the system \eqref{eq:system},
and resolving \eqref{eqn:mpc_ocp} at each subsequent step defines 
the parameterized feedback policy $\hat u_{0 \mid k}^*(x_k; \theta)$.

\subsection{Probabilistic Surrogates}
To systematically optimize the MPC parameters $\theta$ in \eqref{eqn:mpc_ocp}, we require a surrogate model $\hat G : \Theta \to \mathbb{R}$ that probabilistically captures the unknown closed-loop performance function $G$ of the system \eqref{eq:system} under the MPC law $\hat u_{0 \mid k}^*(x_k; \theta)$. Such models provide both predictions and uncertainty estimates, which are essential for data-efficient exploration in Bayesian optimization.

Closed-loop experiments yield (noisy) observations of $G(\theta)$, forming the dataset $\mathcal{D} = \{(\theta_i, y_i) \mid i = 1,\dots,n_{\mathrm{d}}\}$ with $y_i = G(\theta_i) + \varepsilon_i$ and $\varepsilon_i \sim \mathcal{N}(0,\sigma^2)$ representing measurement noise. We collect the evaluated parameters in $\Theta_{\mathrm{d}} \in \mathbb{R}^{n_{\mathrm{d}} \times n_\theta}$ and the corresponding outputs in $y \in \mathbb{R}^{n_{\mathrm{d}}}$, and write $\mathcal{D} = (\Theta_{\mathrm{d}}, y)$ for compactness. A probabilistic surrogate inferred from $\mathcal{D}$ enables performance predictions at new parameters $\theta_*$ together with associated uncertainty estimates. In this work, we investigate Gaussian processes, Bayesian neural networks, and infinite-width BNNs as surrogate models for $G(\theta)$.

\subsubsection{Gaussian Processes}
\label{subsec:GP}
Gaussian process (GP) regression provides a Bayesian nonparametric approach to modeling the surrogate performance function $\hat G$. Loosely speaking, a GP places a Gaussian prior distribution over the space of functions, written as
\[
\hat G(\theta) \sim \mathcal{GP}\big(m(\theta),\,k(\theta,\theta')\big),
\]
where $m : \Theta \to \mathbb{R}$ defines the prior mean and $k : \Theta \times \Theta \to \mathbb{R}$ specifies the covariance structure between function evaluations. Any finite collection of function values follows a joint Gaussian distribution.

Conditioning this prior on the observed closed-loop data $\mathcal{D} = (\Theta_{\mathrm{d}}, y)$ results in the posterior predictive distribution at a new input $\theta_*$. The predictive distribution remains Gaussian and is given by
\[
\hat G(\theta_*) \mid \Theta_{\mathrm{d}}, y, \theta_* \sim \mathcal{N}\!\big(m^+(\theta_*),\,k^+(\theta_*)\big),
\]
with mean and variance computed as
\begin{subequations}
\label{eqn:posterior_gp}
\begin{align}
    m^+(\theta_*) &= m(\theta_*) 
    + k(\theta_*, \Theta_{\mathrm{d}})\,K_y^{-1}
    \big(y - m(\Theta_{\mathrm{d}})\big), \label{eq:gp_postMean} \\
    k^+(\theta_*) &= k(\theta_*,\theta_*)
    - k(\theta_*, \Theta_{\mathrm{d}})\,K_y^{-1}
    k(\Theta_{\mathrm{d}}, \theta_*), \label{eq:gp_postVar}
\end{align}
\end{subequations}
where $K_y = k(\Theta_{\mathrm{d}},\Theta_{\mathrm{d}}) + \sigma^2 I$ and $I$ denotes the identity matrix. The posterior mean~\eqref{eq:gp_postMean} provides a prediction of $G(\theta_*)$, and the posterior variance~\eqref{eq:gp_postVar} reflects the uncertainty in this prediction.

The prior mean $m(\cdot)$, kernel function $k(\cdot,\cdot)$, and their hyperparameters must be selected to reflect structural assumptions about $G(\theta)$. Hyperparameters are typically tuned by maximizing the marginal likelihood of the data $\mathcal{D}$ \citep{rasmussen2006gaussian}.

\subsubsection{Bayesian Neural Networks}
As an aleternativeof surrogates, we consider BNNs \citep{Neal1996}. BNNs retain the overall structure of deterministic feedforward networks, but place a prior over all weights and biases. Conditioning this prior on data yields a posterior distribution over functions $\hat G(\theta;\mathcal{W})$ and hence a predictive distribution with epistemic uncertainty, analogous to the GP posterior.

In contrast to GPs, the computational cost of BNN inference scales primarily with the number of network parameters rather than cubically with the number of data points $n_{\mathrm{d}}$, which makes BNNs attractive for larger datasets or higher-dimensional parameter spaces.

We parameterize the BNN surrogate as $\hat G(\theta;\mathcal{W})$, where $\mathcal{W}$ collects all weights and biases.
A typical choice of prior over $\mathcal{W}$ is an independent Gaussian prior with zero mean and variance~$\nu$ \citep{jospin2022},
\begin{equation}
    p(\mathcal{W}) = \prod_{k} \mathcal{N}\!\big(w_k \mid 0, \nu \big),
\end{equation}
where the product runs over all scalar parameters $w_k \in \mathcal{W}$.
Given the dataset $\mathcal{D} = (\Theta_{\mathrm{d}}, y)$,
we model the likelihood under fixed weights $\mathcal{W}$ as
\begin{equation}
    p(y \mid \Theta_{\mathrm{d}}, \mathcal{W})
    = \prod_{i=1}^{n_{\mathrm{d}}}
      \mathcal{N}\!\big(y_i \mid \hat G(\theta_i;\mathcal{W}), \sigma^2\big).
\end{equation}
Conditioning the prior on $\mathcal{D}$ via Bayes' rule yields the posterior
\begin{equation}
    p(\mathcal{W} \mid \mathcal{D})
    \propto p(y \mid \Theta_{\mathrm{d}}, \mathcal{W})\,p(\mathcal{W}),
    \label{eq:bnn_posterior}
\end{equation}
which is generally intractable due to the nonlinear activations in $\hat G(\cdot;\mathcal{W})$.
Approximate inference methods such as Markov chain Monte Carlo, Laplace approximation, probabilistic backpropagation, or variational inference are therefore used to obtain an approximation $q(\mathcal{W}) \approx p(\mathcal{W} \mid \mathcal{D})$ \citep{abdar2021,daxberger2021}.

Similar to the GP case, the predictive distribution at a new
$\theta_*$ is obtained similarly by marginalizing over this posterior,
\begin{equation*}
   \hat G(\theta_*;\mathcal{W}) \mid \Theta_{\mathrm{d}}, y, \theta_*
    \sim \mathcal{N}\!\big(m_{\mathrm{BNN}}^+(\theta_*),\,
                              k_{\mathrm{BNN}}^+(\theta_*)\big),
\end{equation*}
where the mean and variance are defined as
\begin{subequations}
\label{eqn:posterior_bnn}
\begin{align}
    m_{\mathrm{BNN}}^+(\theta_*)
    &\approx \frac{1}{S}\sum_{s=1}^S \hat G\big(\theta_*;\mathcal{W}^{(s)}\big),
    \\
    k_{\mathrm{BNN}}^+(\theta_*)
    &\approx \frac{1}{S\!-\!1}\sum_{s=1}^S
        \Big(\hat G\big(\theta_*;\mathcal{W}^{(s)}\big)\!-\!m_{\mathrm{BNN}}^+(\theta_*)\Big)^2,
\end{align}
\end{subequations}
with Monte Carlo samples $\mathcal{W}^{(s)} \sim q(\mathcal{W}) \approx
p(\mathcal{W}\mid\mathcal{D})$, $s = 1,\dots,S$.

\subsubsection{Infinite-Width Bayesian Neural Networks}
Infinite-width Bayesian neural networks (I-BNNs) are the result of increasing the number of units in each hidden layer of a BNN, and as a consequence, the number of weights and biases towards infinity.
In this limit, the prior over functions $\hat G(\theta)$ induced by the network converges to a Gaussian process~\citep{Neal1996,lee2018deep}.

Consider a depth-$L$ fully connected network with scalar output and input $\theta$. For a layer $\ell \in [1,\hdots,L]$, with $n_{\ell-1}$ inputs, we place independent Gaussian priors
\begin{equation}
   W^{(\ell)}_{ij} \sim \mathcal{N}\!\Big(0,\frac{\sigma_w^2}{n_{\ell-1}}\Big), 
   \qquad
   b^{(\ell)}_j \sim \mathcal{N}(0,\sigma_b^2),
\end{equation}
over weights and biases, with $\sigma_w^2$ and $\sigma_b^2$ denoting their prior variances. Under this scaling, as $n_{\ell-1} \to \infty$ for all hidden layers, the output of the network at any finite collection of inputs becomes jointly Gaussian, and hence
\begin{equation}
    \hat G(\theta) \sim \mathcal{GP}\big(0,\,k_{\mathrm{IBNN}}(\theta,\theta')\big),
\end{equation}
where the prior mean is zero due to the zero-mean weight and bias priors, and the kernel $k_{\mathrm{IBNN}}$ is determined by the network architecture, activation functions, prior variances $\sigma_w^2$ and $\sigma_b^2$, and depth $L$.

The kernel can be obtained by propagating covariances layer by layer. Let
\begin{equation}
    K^{0}(\theta,\theta') 
    = \sigma_b^2 + \frac{\sigma_w^2}{n_\theta}\,\theta^\top \theta'
\end{equation}
denote the covariance induced by the linear input layer. We define the normalized correlation
\begin{equation}
    \rho^{\ell-1}(\theta,\theta')
    = \frac{K^{\ell-1}(\theta,\theta')}
           {\sqrt{K^{\ell-1}(\theta,\theta)\,K^{\ell-1}(\theta',\theta')}}.
\end{equation}
With elementwise activation function $\phi$, the covariance at layer $\ell$ is then given by
\begin{equation}
    K^{\ell}(\theta,\theta')
    = \sigma_b^2 + \sigma_w^2\,
      \mathbb{E}_{(z_1,z_2)\sim \mathcal{N}
        \left(0, C^{\ell -1}(\theta,\theta')\right)}
      \big[\phi(z_1)\,\phi(z_2)\big],
\end{equation}
where $(z_1,z_2)^\top$ denotes a zero-mean bivariate Gaussian random vector with correlation matrix
\begin{equation*}
    C^{\ell-1}(\theta,\theta') =
    \begin{bmatrix}
        1 & \rho^{\ell-1}(\theta,\theta') \\
        \rho^{\ell-1}(\theta,\theta') & 1
    \end{bmatrix}.
\end{equation*}
For common choices of $\phi$ (e.g., ReLU or $\tanh$), this expectation admits a closed-form expression~\citep{lee2018deep}. The resulting I-BNN kernel is given by $k_{\mathrm{IBNN}}(\theta,\theta') = K^{L}(\theta,\theta')$ at the chosen depth $L$.

In practice, I-BNNs are GPs and can be used for BO analog to the GP case, see Section~\ref{subsec:GP}, differing only in the choice of kernel.
Compared to finite-width BNNs, the infinite-width limit yields an analytically tractable GP posterior and removes the need for sampling-based approximate inference.

\subsection{Bayesian Optimization}
Having introduced probabilistic surrogates for modeling the closed-loop cost, we now employ these models within a Bayesian optimization (BO) framework to iteratively search for MPC parameters that maximize performance. Since the closed-loop cost can only be accessed through computationally or experimentally expensive closed-loop evaluations, direct global optimization is impractical, motivating the use of a data-efficient sequential strategy such as BO \citep{garnett2023bayesian}.

Our objective is to solve the closed-loop learning problem \ref{eq:learning_problem} for the optimal parameters $\theta^*$, where the surrogate model provides a probabilistic approximation of the unknown function $G$. BO proceeds by alternating between selecting a new parameter point and updating the surrogate based on the resulting observation. In iteration $n \in \mathbb{N}$, the following steps are performed:
\begin{enumerate}
    \item[1)] select a candidate parameter vector $\theta_n$ based on the current surrogate,
    \item[2)] evaluate the closed-loop performance $G(\theta_n)$ to obtain a new noisy data point,
    \item[3)] update the dataset $\mathcal{D}_{n+1} = \mathcal{D}_n \cup \{(\theta_n, G(\theta_n))\}$ and refine the surrogate model using $\mathcal{D}_{n+1}$.
\end{enumerate}

The decision which parameter $\theta_n$ to evaluate next is guided by an acquisition function $\alpha(\theta; \mathcal{D}_n)$ that quantifies the expected utility of sampling at $\theta$ given the current model of $G$. Leveraging both predictive mean and uncertainty, BO balances exploring insufficiently sampled regions and exploiting promising parameter areas. The next query point is chosen by maximizing the acquisition function according to
\begin{equation}
    \label{eqn:bo_update}
    \theta_{n+1} = \arg \max_{\theta \in \Theta} \alpha(\theta; \mathcal{D}_n).
\end{equation}
This iterative scheme progressively improves the estimate of $\theta^*$ while keeping the number of closed-loop evaluations low.

\begin{figure*}
    \includegraphics[width=0.99\textwidth]{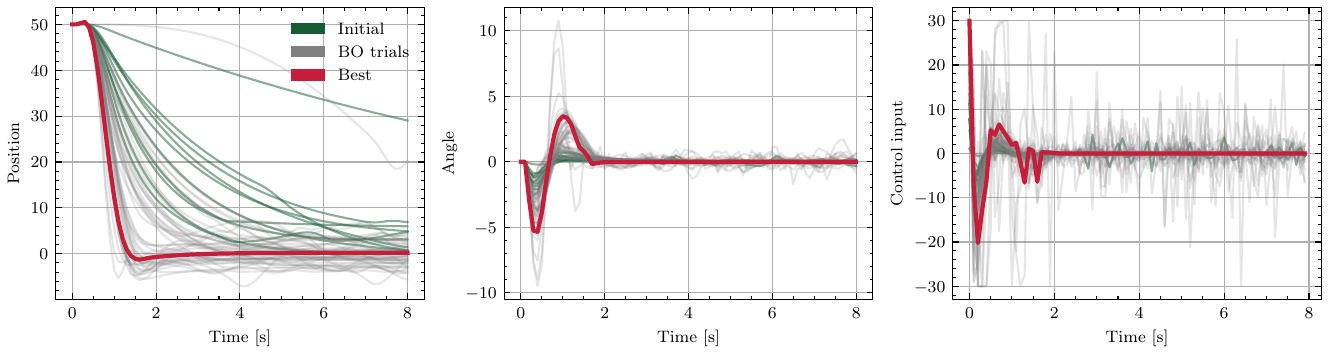}
    \caption{Closed-loop state trajectories and closed-loop control inputs from a single BO run using an I-BNN surrogate. Initial random parameter samples (green), intermediate BO-driven rollouts (gray), and the best-performing trajectory (red) illustrate the improvement in closed-loop performance achieved through high-dimensional neural-cost learning.}
    \label{fig:closed_loop}
\end{figure*}

\section{Neural Cost Function Parameterization}
\label{sec:nn_param}
We are interested in neural network parameterizations of the MPC cost to directly optimize the high-level closed-loop performance from data, as they offer high flexibility in formulation. Even the parameters of classical MPC stage costs with quadratic structure can be challenging to learn effectively, and their fixed parametric form limits achievable closed-loop behavior. By augmenting the cost with a neural network term, we enable a more flexible and expressive parameterization, leading to improved closed-loop performance. However, this results in a high-dimensional parameter space due to the many neural network parameters that need to be learned. This provides a suitable setting to study how different probabilistic surrogate models, such as GPs, BNNs, and I-BNNs, scale as surrogate models within Bayesian optimization for controller tuning in high-dimensional parameter spaces.

The first part of the stage cost follows the standard quadratic form
\begin{equation}
    \label{eqn:quadratic_state_cost}
    l_q(x,u;\theta) = (x-x_d)^\top Q_\theta (x-x_d) + (u-u_d)^\top R_\theta (u-u_d),
\end{equation}
where $Q_\theta = \operatorname{diag}(q_\theta)$ and $R_\theta = \operatorname{diag}(r_\theta)$ with $q_\theta \in \mathbb{R}^{n_{\text{x}}}_{\ge 0}$ and $r_\theta \in \mathbb{R}^{n_{\text{u}}}_{>0}$ denoting parameterized diagonal matrices satisfying $Q_\theta \succeq 0$ and $R_\theta \succ 0$.

To enhance expressiveness, we introduce the parameterized stage cost
\begin{equation}
    \label{eqn:specific_stage_cost}
    l_\theta(x, u) = l_q(x,u; \theta) + l_{NN}(x; \theta),
\end{equation}
where $\theta$ denotes all trainable parameters. To preserve the control objective, we enforce
\begin{equation}\label{eq:NN_cost_term}
    l_{NN}(x; \theta) = y_{NN}(x; \theta)-y_{NN}(x_d; \theta),
\end{equation}
with $y_{NN}: \mathbb{R}^{n_{\text{x}}} \to \mathbb{R}$ defined as a feedforward neural network with $L$ layers,
\begin{equation*}
    y_{NN}(z; \theta) = W_L \chi(W_{L-1} \chi(...\chi(W_1 z + b_1)... ) + b_{L-1}) + b_L.
\end{equation*}
Here, $\chi$ denotes the activation function, and $(W_i,b_i)$ are the weights and biases of layer $i$, $i=1,\dots,L$.

Consequently, the full parameter vector optimized during learning is
\begin{equation*}
\theta = \big[\, q_\theta^\top,\ r_\theta^\top,\ \mathrm{vec}(W_1)^\top,\ b_1^\top,\ \dots,\ \mathrm{vec}(W_L)^\top,\ b_L^\top \,\big]^\top,
\end{equation*}
where $\mathrm{vec}(\cdot)$ denotes vectorization of a matrix, such that all scalar entries are stacked into the single parameter vector $\theta \in \mathbb{R}^{n_\theta}$. 
This results in a high-dimensional search space, motivating the use of sample-efficient global optimization techniques such as Bayesian optimization for learning the cost function parameters.

\section{Efficiency of different surrogate models via simulations}
\label{sec:simulation}
We examine the effectiveness of different surrogates for high-dimensional parameter learning in the following in simulation. After describing the setup, we present closed-loop performance results under different MPC cost parameterizations and evaluate the different surrogate models within an Bayesian optimization routine.

\subsection{Set-Up}
We evaluate the proposed Bayesian optimization framework on a linearized cart–pole system with state vector $x = [x_\mathrm{c}, \dot{x}_\mathrm{c}, \varphi, \dot{\varphi}]^\top \in \mathbb{R}^4$ and control input $u \in \mathbb{R}$ being the cart acceleration. The linearized dynamics are discretized and used directly within MPC with prediction horizon $N = 10$ and state and input box constraints. Performance is quantified by the closed-loop cost over an evaluation horizon $M = 80$,
\begin{align*}
    G(\theta) =& \sum_{k=0}^{M} \left(x_k(\theta) - x_\mathrm{d}\right)^\top Q_\mathrm{BO} \left(x_k(\theta) - x_\mathrm{d}\right) \\
     & + \left(u_k(\theta) - u_\mathrm{d}\right)^\top R_\mathrm{BO} \left(u_k(\theta) - u_\mathrm{d}\right) \\
     + &\sum_{k=\bar M}^{M} \left(x_k(\theta) - x_\mathrm{d}\right)^\top P_\mathrm{BO} \left(x_k(\theta) - x_\mathrm{d}\right)
\end{align*}
Here, $\bar M = 70$, $Q_\mathrm{BO} = \mathrm{diag}(1,\,0.1,\,0.1,\,0.1)$, $R_\mathrm{BO} = 1$, and $P_\mathrm{BO} = \mathrm{diag}(100,\,10,\,10,\,10)$ are fixed evaluation weight matrices used only for assessing closed-loop performance and are distinct from the MPC weighting matrices $Q_\theta$ and $R_\theta$. The cost is accumulated over the entire episode, with an evaluation horizon $M \gg N$, to reflect the long-term closed-loop behavior of the controller. The additional sum from $\bar M$ to $M$ increases the penalty on the tail of the episode to additionally promote convergence of the closed-loop trajectory toward the desired state $x_\mathrm{d}$.

To investigate scalability with respect to the dimensionality of the MPC cost parameterization, we vary the number of neurons in the neural network stage cost while always including the diagonal entries of $Q_\theta$ and $R_\theta$ in the learned parameters. This results in settings ranging from low- to high-dimensional controller parameter spaces. Across all settings, only the surrogate model in BO is varied, while all other components remain identical to ensure a fair comparison between methods. We use the same acquisition function, the logarithmic Expected Improvement, as well as identical optimization routines, search bounds, and initial training data. A fixed number of initial design points is evaluated using random parameter samples.

We compare three different surrogate models: Gaussian processes with a Matérn kernel, which represents the standard baseline in the BO literature, BNNs, and infinite-width BNNs. Additionally, we consider sequential random sampling of parameters as a baseline. For each configuration, multiple BO runs with different random seeds are conducted. Since the true optimal parametrization is unknown, the true regret cannot be computed; therefore, we evaluate performance by tracking the best observed closed-loop cost over BO iterations.

\subsection{BNNs and I-BNNs as alternative to Gaussian Processes}
Figure~\ref{fig:closed_loop} illustrates the influence of neural cost learning on the closed-loop cart--pole behavior for a single BO run with an I-BNN surrogate. Initial random samples lead to widely spread trajectories with pronounced oscillations and unstructured input profiles. As BO progresses, trajectories concentrate around better behaviors, and the best-found parameterization achieves faster convergence to the desired state, reduced oscillations, and a more structured control input.

\begin{figure}
    \begin{minipage}[b!]{0.47\textwidth}
        \includegraphics[width=\textwidth]{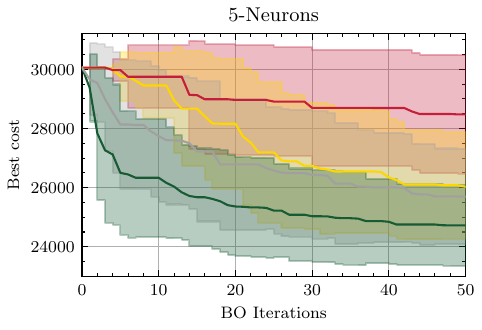}
    \end{minipage}
    \begin{minipage}[b!]{0.47\textwidth}
        \vspace{-9mm}
        \includegraphics[width=\textwidth]{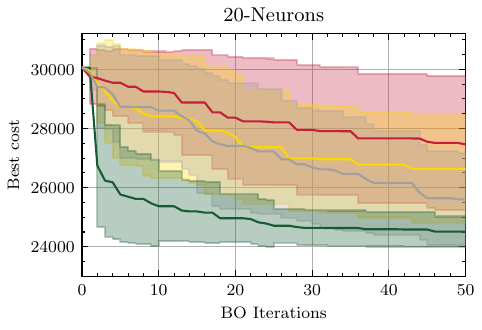}
    \end{minipage}
    \begin{minipage}[b!]{0.47\textwidth}
        \vspace{-9mm}
        \includegraphics[width=\textwidth]{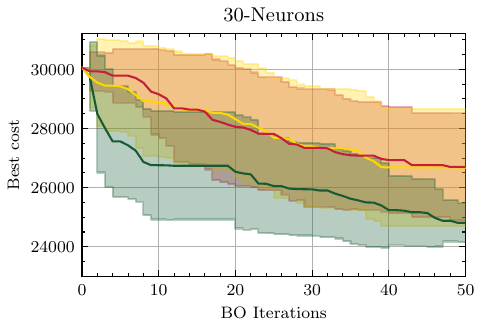}
    \end{minipage}
    \vspace{-2mm}
    \caption{Best observed closed-loop cost over BO iterations for neural-network cost parameterizations of 5 neurons (66 parameters), 20 neurons (546 parameters), and 30 neurons (1116 parameters). Curves show the mean and one standard deviation over 21 independent BO runs each. Red indicates random sampling, yellow the Matérn-kernel GP surrogate, silver the finite-width BNN surrogate, and green the infinite-width BNN surrogate.}
    \label{fig:regret}
\end{figure}

Figure~\ref{fig:regret} quantifies the impact of different surrogate models on the convergence of the best observed closed-loop cost for MPC cost networks with two hidden layers and $5$, $20$, and $30$ neurons per hidden layer, corresponding to $66$, $546$, and $1116$ parameters, respectively.
For $5$ neurons, all BO-based surrogates (Matérn GP, finite-width BNN, and I-BNN) clearly outperform random search, with I-BNN showing a slight advantage in convergence speed and final cost. For $20$ neurons, performance differences become pronounced: I-BNN achieves the fastest and lowest cost convergence, finite-width BNN still clearly improves over Matérn GP and random search, while the Matérn GP behaves only marginally better than random. For $30$ neurons, we compare only random, Matérn GP, and I-BNN, since MCMC-based finite-width BNNs are computationally prohibitive in this regime. All methods degrade with increasing dimension, but the Matérn GP is effectively indistinguishable from random sampling in the case of $1116$ parameters, whereas I-BNN still provides a visible reduction in cost and a lower final cost.

Overall, the results show that BNN-based surrogates, and I-BNNs in particular, scale more favorably with the dimensionality of dense NN-based MPC cost parameterizations than Matérn GPs. While standard Matérn GPs remain a simple and viable choice in low dimensions, they are particularly attractive there due to their cheap kernel evaluations and simple hyperparameter structure. However, their modeling fidelity quickly deteriorates once the parameter space reaches a few hundred dimensions, and they essentially collapse to random-like behavior beyond $1000$ parameters. Finite-width BNNs offer a useful intermediate option up to a few hundred parameters but become too expensive at very high dimensionality due to MCMC-based inference. In contrast, I-BNN surrogates retain a clear performance advantage even in the $1116$-dimensional case and therefore appear as the most promising choice for high-dimensional neural cost tuning in BO-based MPC.

\section{Conclusion}
\label{sec:conclusion}
Motivated by the difficulty of applying Bayesian optimization to dense high-dimensional controller parameterizations, we investigated probabilistic surrogate models for learning neural-network–parameterized MPC cost functions. We considered standard Matérn-kernel Gaussian processes, finite-width Bayesian neural networks, and infinite-width Bayesian neural networks on a cart–pole system with up to 1116 cost parameters, and evaluated their closed-loop performance in terms of convergence of the closed-loop cost and closed-loop behavior.

Our simulation results suggest that Bayesian-neural-network-based surrogate models, and infinite-width Bayesian neural networks in particular, scale more favorably than Matérn-kernel Gaussian processes in dense high-dimensional parameter spaces. While all surrogate models perform well in low dimensions, Matérn-kernel Gaussian processes lose effectiveness as the number of controller parameters grows and approach random-search behavior beyond a few hundred dimensions. Finite-width Bayesian neural networks offer strong performance up to this range but become computationally prohibitive in very high dimensions. Infinite-width Bayesian neural networks, by contrast, maintain learning performance even for cost networks with more than 1000 parameters. These observations provide practical guidance for surrogate model selection in learning-based MPC: Matérn-kernel Gaussian processes remain computationally attractive and effective in low-dimensional settings, but as dimensionality increases, their expressiveness becomes insufficient, whereas infinite-width Bayesian neural networks may remain suitable for dense high-dimensional parameterizations.

This work focuses on a single dynamical system and a specific neural cost structure, providing a clear setting to assess the scalability of different surrogate models. Future work will extend the evaluation to additional controller parameterizations, dynamical systems, experimental setups, and the integration of infinite-width Bayesian neural network surrogates with stability and safety considerations.

\begin{ack}
This research was supported by the German Research Foundation (DFG) within RTG 2761 LokoAssist under grant no. 450821862.
\end{ack}
\section*{DECLARATION OF GENERATIVE AI AND AI-ASSISTED TECHNOLOGIES IN THE WRITING PROCESS}
During the preparation of this work the author(s) used ChatGPT in order to improve spelling and grammar solely. After using this tool/service, the author(s) reviewed and edited the content as needed and take(s) full responsibility for the content of the publication.
\bibliography{ifacconf}

\begin{thebibliography}{22}
\providecommand{\natexlab}[1]{#1}
\providecommand{\url}[1]{\texttt{#1}}
\providecommand{\urlprefix}{URL }
\expandafter\ifx\csname urlstyle\endcsname\relax
  \providecommand{\doi}[1]{doi:\discretionary{}{}{}#1}\else
  \providecommand{\doi}{doi:\discretionary{}{}{}\begingroup \urlstyle{rm}\Url}\fi

\bibitem[{Abdar et~al.(2021)Abdar, Pourpanah, Hussain, Rezazadegan, Liu, Ghavamzadeh, Fieguth, Cao, Khosravi, Acharya, Makarenkov, and Nahavandi}]{abdar2021}
Abdar, M., Pourpanah, F., Hussain, S., Rezazadegan, D., Liu, L., et~al. (2021).
\newblock {A review of uncertainty quantification in deep learning: Techniques, applications and challenges}.
\newblock \emph{Information Fusion}, 76, 243--297.
\newblock \doi{10.1016/j.inffus.2021.05.008}.

\bibitem[{Daxberger et~al.(2021)Daxberger, Kristiadi, Immer, Eschenhagen, Bauer, and Hennig}]{daxberger2021}
Daxberger, E., Kristiadi, A., Immer, A., Eschenhagen, R., Bauer, M., et~al. (2021).
\newblock Laplace redux --- effortless {{Bayesian}} deep learning.
\newblock In \emph{Proce. Int. Conf. Neural Inf. Proc. Syst.} NIPS.

\bibitem[{Eriksson and Jankowiak(2021)}]{eriksson2021high-dimensional}
Eriksson, D. and Jankowiak, M. (2021).
\newblock High-dimensional {{Bayesian}} optimization with sparse axis-aligned subspaces.
\newblock In C.~de~Campos and M.H. Maathuis (eds.), \emph{Proceedings of the Thirty-Seventh Conference on Uncertainty in Artificial Intelligence}, volume 161 of \emph{Proceedings of Machine Learning Research}, 493--503. PMLR.

\bibitem[{Findeisen et~al.(2026)Findeisen, Rose, Graichen, and Mönnigmann}]{Findeisen2025}
Findeisen, R., Rose, A., Graichen, K., and Mönnigmann, M. (2026).
\newblock Embedded optimization in control: An introduction, opportunities, and challenges.
\newblock In Z.~Ding (ed.), \emph{Encyclopedia of Systems and Control Engineering (First Edition)}, 223--244. Elsevier, Oxford.

\bibitem[{Garnett(2023)}]{garnett2023bayesian}
Garnett, R. (2023).
\newblock \emph{Bayesian Optimization}.
\newblock {Cambridge University Press}.

\bibitem[{Hirt et~al.(2024{\natexlab{a}})Hirt, Höhl, Schaeffer, Pohlodek, Braatz, and Findeisen}]{hirt2025learning}
Hirt, S., Höhl, A., Schaeffer, J., Pohlodek, J., Braatz, R.D., et~al. (2024{\natexlab{a}}).
\newblock Learning model predictive control parameters via {{Bayesian}} optimization for battery fast charging.
\newblock \emph{IFAC-PapersOnLine}, 58(14), 742--747.
\newblock 12th IFAC Symposium on Advanced Control of Chemical Processes ADCHEM 2024.

\bibitem[{Hirt et~al.(2024{\natexlab{b}})Hirt, Pfefferkorn, and Findeisen}]{hirt2024safe}
Hirt, S., Pfefferkorn, M., and Findeisen, R. (2024{\natexlab{b}}).
\newblock Safe and stable closed-loop learning for neural-network-supported model predictive control.
\newblock In \emph{2024 IEEE 63rd Conference on Decision and Control (CDC)}, 4764--4770.

\bibitem[{Hirt et~al.(2024{\natexlab{c}})Hirt, Pfefferkorn, Mesbah, and Findeisen}]{hirt2024stabilityinformed}
Hirt, S., Pfefferkorn, M., Mesbah, A., and Findeisen, R. (2024{\natexlab{c}}).
\newblock Stability-informed {{Bayesian}} optimization for {{MPC}} cost function learning.
\newblock \emph{IFAC-PapersOnLine}, 58(18), 208--213.

\bibitem[{Jospin et~al.(2022)Jospin, Laga, Boussaid, Buntine, and Bennamoun}]{jospin2022}
Jospin, L.V., Laga, H., Boussaid, F., Buntine, W., and Bennamoun, M. (2022).
\newblock Hands-on {{Bayesian}} neural networks --- {{A}} tutorial for deep learning users.
\newblock \emph{IEEE Comp. Int. Mag.}, 17(2), 29--48.
\newblock \doi{10.1109/MCI.2022.3155327}.

\bibitem[{Kordabad et~al.(2023)Kordabad, Reinhardt, Anand, and Gros}]{kordabad2023reinforcement}
Kordabad, A.B., Reinhardt, D., Anand, A.S., and Gros, S. (2023).
\newblock Reinforcement learning for {{MPC}}: Fundamentals and current challenges.
\newblock In \emph{IFAC-PapersOnLine}, volume~56, 5773--5780.

\bibitem[{Lee et~al.(2018)Lee, Bahri, Novak, Schoenholz, Pennington, and Sohl-Dickstein}]{lee2018deep}
Lee, J., Bahri, Y., Novak, R., Schoenholz, S.S., Pennington, J., et~al. (2018).
\newblock Deep neural networks as {{Gaussian}} processes.
\newblock In \emph{International Conference on Learning Representations}.

\bibitem[{Li et~al.(2024)Li, Rudner, and Wilson}]{li2024study}
Li, Y.L., Rudner, T.G.J., and Wilson, A.G. (2024).
\newblock A study of {{Bayesian}} neural network surrogates for {{Bayesian}} optimization.
\newblock \doi{10.48550/arXiv.2305.20028}.

\bibitem[{Lucia et~al.(2016)Lucia, K{\"o}gel, Zometa, Quevedo, and Findeisen}]{lucia2016predictive}
Lucia, S., K{\"o}gel, M., Zometa, P., Quevedo, D.E., and Findeisen, R. (2016).
\newblock Predictive control, embedded cyberphysical systems and systems of systems--a perspective.
\newblock \emph{Annual Reviews in Control}, 41, 193--207.

\bibitem[{Makrygiorgos et~al.(2022)Makrygiorgos, Bonzanini, Miller, and Mesbah}]{makrygiorgos2022performanceorienteda}
Makrygiorgos, G., Bonzanini, A.D., Miller, V., and Mesbah, A. (2022).
\newblock Performance-oriented model learning for control via multi-objective {{Bayesian}} optimization.
\newblock \emph{Computers \& Chemical Engineering}, 162, 107770.

\bibitem[{Moss et~al.(2025)Moss, Ober, and Diethe}]{moss2025return}
Moss, H.B., Ober, S.W., and Diethe, T. (2025).
\newblock Return of the latent space cowboys: Re-thinking the use of {{VAEs}} for {{Bayesian}} optimisation of structured spaces.
\newblock \emph{arXiv preprint arXiv:2507.03910}.

\bibitem[{Neal(1996)}]{Neal1996}
Neal, R.M. (1996).
\newblock \emph{Bayesian Learning for Neural Networks}.
\newblock Lecture Notes in Statistics. Springer.

\bibitem[{Paulson et~al.(2023)Paulson, Sorourifar, and Mesbah}]{paulson2023tutorial}
Paulson, J.A., Sorourifar, F., and Mesbah, A. (2023).
\newblock A tutorial on derivative-free policy learning methods for interpretable controller representations.
\newblock In \emph{American Control Conference}, 1295--1306.

\bibitem[{Piga et~al.(2019)Piga, Forgione, Formentin, and Bemporad}]{piga2019performance}
Piga, D., Forgione, M., Formentin, S., and Bemporad, A. (2019).
\newblock Performance-oriented model learning for data-driven {{MPC}} design.
\newblock \emph{IEEE Control Systems Letters}, 3(3), 577--582.

\bibitem[{Rasmussen and Williams(2006)}]{rasmussen2006gaussian}
Rasmussen, C.E. and Williams, C.K. (2006).
\newblock \emph{Gaussian Processes for Machine Learning}.
\newblock {MIT Press, Cambridge, MA}.

\bibitem[{Rawlings et~al.(2017)Rawlings, Mayne, and Diehl}]{rawlings2017model}
Rawlings, J.B., Mayne, D.Q., and Diehl, M. (2017).
\newblock \emph{Model Predictive Control: {{Theory}}, Computation, and Design}.
\newblock {Nob Hill Publishing}, {Madison, Wisconsin}, 2nd edition.

\bibitem[{Schwenzer et~al.(2021)Schwenzer, Ay, Bergs, and Abel}]{schwenzer2021review}
Schwenzer, M., Ay, M., Bergs, T., and Abel, D. (2021).
\newblock Review on model predictive control: An engineering perspective.
\newblock \emph{The International Journal of Advanced Manufacturing Technology}, 117(5), 1327--1349.

\bibitem[{Seel et~al.(2022)Seel, Kordabad, Gros, and Gravdahl}]{seel2022convex}
Seel, K., Kordabad, A.B., Gros, S., and Gravdahl, J.T. (2022).
\newblock Convex neural network-based cost modifications for learning model predictive control.
\newblock \emph{IEEE Open Journal of Control Systems}, 1, 366--379.

\end{thebibliography}

\end{document}